\documentclass[11pt,a4paper]{article}
\usepackage[hyperref]{emnlp2020}
\usepackage{graphicx}
\graphicspath{ {./images/} }

\usepackage{pgfplots}
\usepackage{pgfplotstable}
\usetikzlibrary[patterns]

\usepackage{times}
\usepackage{latexsym}

\usepackage{microtype}

\usepackage{booktabs}
\usepackage{multirow}
\usepackage{adjustbox}

\usepackage{soul} 

\newcommand{\dir}{\texttt{dir}}
\newcommand{\ddir}{\texttt{2 dir}}
\newcommand{\dddir}{\texttt{3 dir}}
\newcommand{\dirl}{\texttt{dir + lm}}

\newcommand{\dddirl}{\texttt{3 dir + lm}}
\newcommand{\dirlc}{\texttt{dir + lm + ch}}

\newcommand{\cmbig}{\texttt{big}}
\newcommand{\cmbase}{\texttt{base}}
\newcommand{\cmhalf}{\texttt{half}}

\newcommand{\cmsixteenth}{\texttt{16th}}

\newcommand{\cmbigs}{\texttt{big\_1\_1}}
\newcommand{\cmbases}{\texttt{base\_1\_1}}

\newcommand{\cmsixteenths}{\texttt{16th\_1\_1}}

\aclfinalcopy 
\def\aclpaperid{43} 

\title{Language Models not just for Pre-training:\\Fast Online Neural Noisy Channel Modeling}

\author{Shruti Bhosale$^{\bigtriangleup}$ \quad Kyra Yee$^{\dagger\bigtriangledown}$ \quad Sergey Edunov$^{\bigtriangleup}$ \quad Michael Auli$^{\bigtriangleup}$ \\
  $^{\bigtriangleup}$ Facebook AI Research, Menlo Park, CA, USA \\
  $^{\bigtriangledown}$ Twitter Cortex, San Francisco, CA, USA \\
}

\begin{document}
\maketitle

\renewcommand* {\thefootnote}{\fnsymbol{footnote}}
\footnotetext{$\dagger$ Work done while at Facebook during a Facebook AI Residency.}
\renewcommand*{\thefootnote}{\arabic{footnote}}

\interfootnotelinepenalty=30000

\begin{abstract}
Pre-training models on vast quantities of unlabeled data has emerged as an effective approach to improving accuracy on many NLP tasks.
On the other hand, traditional machine translation has a long history of leveraging unlabeled data through noisy channel modeling. 
The same idea has recently been shown to achieve strong improvements for neural machine translation.
Unfortunately, na\"{i}ve noisy channel modeling with modern sequence to sequence models is up to an order of magnitude slower than alternatives. 
We address this issue by introducing efficient approximations to make inference with the noisy channel approach as fast as strong ensembles while increasing accuracy.
We also show that the noisy channel approach can outperform strong pre-training results by achieving a new state of the art on WMT Romanian-English translation. 
\end{abstract}

\section{Introduction}

Unlabeled data has been leveraged in many ways in natural language processing including  back-translation~\citep{bojar2011bt,sennrich2016bt,edunov2018bt}, self-training~\citep{he2019revisiting}, or language model pre-training which led to improvements in many natural language tasks~\citep{devlin2019bert}.
While pre-training has achieved impressive results on tasks where labeled data is limited, improvements in settings with abundant labeled data are modest~\citep{raffel2020exploring} with controlled studies showing a clear trend of diminishing returns as the amount of training data increases~\citep{edunov2019lmpretrain}.

In this paper, we focus on noisy channel modeling for text generation tasks, a classical technique from the statistical machine translation literature which had been the workhorse of text generation tasks for decades before the arrival of neural sequence to sequence models~\citep{brown1993mathematics,koehn2003statistical}.
Unlike pre-training approaches, this approach is very effective irrespective of the amount of labeled data:
since a recent revival~\citep{yu2017neuralnoisy,yee2019noisy}, it has been an important part in the winning entries of several high resource language pairs at WMT 2019~\citep{ng2019fairwmt}, improving over strong ensembles that used 500M back-translated sentences. 
At the low resource WAT 2019 machine translation competition, noisy channel modeling was also a key factor for the winning entry~\citep{chen2019facebook}.

Noisy channel modeling turns text generation on the head: instead of modeling an output sequence given an input, Bayes' rule is applied to model the input given the output, via a backward sequence to sequence model which is combined with the prior probability of the output, typically a language model. 
This enables the effective use of strong language models trained on large amounts of unlabeled data. 
The role of the backward model, or the channel model, is to validate outputs preferred by the language model with respect to the input.

A straightforward way to use language models is to pair them with standard sequence to sequence models~\citep{gulcehre2015mono,stahlberg2018simple}.
However, this does not address \emph{explaining away effects} under which modern neural sequence models still suffer~\citep{klein2001explain,li2019dont}.
As a consequence, models are susceptible to producing fluent outputs that are unrelated to the input~\citep{li2019dont}.
The noisy channel approach explicitly addresses this via the channel model. 

However, a major obstacle to efficient noisy channel modeling is that generating outputs is much slower than decoding from a standard sequence to sequence model.
We address this issue by introducing several simple yet highly effective approximations which increase the speed of noisy channel modeling by an order of magnitude to make it practical.
This includes smaller channel models as well as scoring only a subset of the channel model vocabulary. 
Experiments on WMT English-Romanian translation show that noisy channel modeling can outperform recent pre-training results.
Moreover, we show that noisy channel modeling benefits much more from larger beam sizes than strong pre-training methods.

\section{The Noisy Channel Approach}
\label{sec:ncapproach}
We assume a sequence to sequence task that takes the input $x$ to predict the output $y$.
A standard sequence to sequence model directly estimates the probability $p(y|x)$, referred to as a \emph{direct model}.
On the other hand, the noisy channel approach applies Bayes' rule to model $p(y|x) = p(x|y) p(y) / p(x)$ where $p(x|y)$ predicts the source $x$ given the target $y$ and is referred to as the \emph{channel model}, $p(y)$ is a language model over the target $y$, and $p(x)$ is generally not modeled since it is constant for all $y$.

\citet{yee2019noisy} use Transformer models to parameterize the direct model, the channel model and the language model. 
Similar to~\citet{yu2017neuralnoisy}, they use the following linear combination of the channel model, the language model as well as the direct model for decoding:
\begin{equation}
\frac{1}{t}\log p(y|x) + \frac{\lambda_1}{s} \log p(x|y) + \frac{\lambda_2}{s} \log p(y)
\label{eq:comb}
\end{equation}
where $t$ is the length of the output prefix $y$, $s$ is the length of the input sequence, and $\lambda_1$, $\lambda_2$ are hyperparameters.

Exact noisy channel model scoring with neural networks during decoding is prohibitively expensive since it requires a separate forward computation with the channel model for every token in the target vocabulary.
To side step this issue, \citet{yu2017neuralnoisy} propose the following approximations to beam search with beam width $k_1$: 
determine the $k_2$ highest scoring extensions of each beam according to the direct model, then score the resulting $k_1 \times k_2$ partial candidates by the direct model, the channel model and the language model using the linear combination in~\autoref{eq:comb}. 
Finally, this set is pruned to beam size $k_1$.

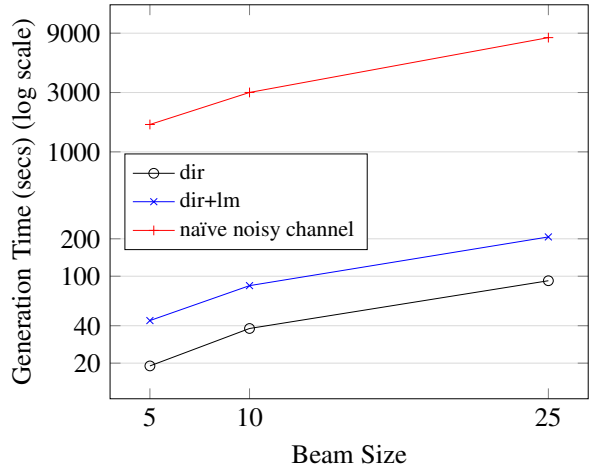
\begin{figure}
\centering
\begin{adjustbox}{width=\columnwidth}
\begin{tikzpicture}
\pgfplotsset{
    grid style={lightgray!60!white}
}
\begin{axis}[
    legend style={font=\footnotesize},
    legend cell align={left},
    xmode=linear,
    xtick={5,10,25},
    xticklabels={5,10,25},
    ytick={20,40,100,200,1000,3000,9000},
    yticklabels={20,40,100,200,1000,3000,9000},
    xlabel=Beam Size,
    ylabel=Generation Time (secs) (log scale),
    ymode=log,
    legend style={at={(0.03,0.5)},anchor=west},
    ymajorgrids
];
\addplot[black, mark=o] table [y=dir, x=type]{data/bad_gen_times_best_bsz.dat};
\addplot[blue, mark=x] table [y=dir+lm, x=type]{data/bad_gen_times_best_bsz.dat};
\addplot[red, mark=+] table [y=dir+ch+lm, x=type]{data/bad_gen_times_best_bsz.dat};
\addlegendentry{dir};
\addlegendentry{dir+lm};
\addlegendentry{na\"{i}ve noisy channel};
\end{axis}
\end{tikzpicture}
\end{adjustbox}
\caption{Speed of decoding with a direct model (\dir{}), direct model with language model (\dirl{}) and a na\"{i}ve noisy channel approach without fast approximations or optimizations. 
The latter is very slow compared to the direct model. Results are based on generation with the fastest batch size for each setting with beam 5 on newstest2016 De-En (cf. \textsection\ref{sec:datasets}).}
\label{fig:slow_noisy}
\end{figure}

Despite this approximation, noisy channel decoding is still significantly slower than decoding with the direct model alone as shown in~\autoref{fig:slow_noisy}. 
The reason for this is that the channel model repeatedly scores the entire input sequence at each time-step and this is done $k_2$ times for each beam.
Specifically, both the direct model and the language model compute $k_1 \times V$ scores at each time-step in order to make a decoding decision for each target token during beam search, where $V$ denotes the vocabulary size which we assume to be similar between the input and output.
In contrast, the channel model computes $k_1 \times k_2 \times S \times V$ scores for each target token, where $S$ is the maximum source sequence length. This adds substantial compute and memory overhead, to the extent that the batch size at decoding often needs to be substantially reduced. This leads to slower inference on GPUs since less computation can be parallelized.

\section{Fast Noisy Channel Modeling}
\label{sec:fastnc}

Na\"{i}ve online noisy channel modeling is significantly slower than standard direct models.
In this section, we present approximations to make noisy channel modeling substantially faster.

\subsection{Reducing Channel Model Size}
\label{sec:smallch}

Prior work on neural noisy channel used channel models which were of the same size as the direct model~\citep{yu2017neuralnoisy,yee2019noisy}. 
The most recent work uses standard Transformer models~\citep{yee2019noisy,ng2019fairwmt,yu2020tacl}.
In this study, we hypothesize that the primary role of the channel model is to avoid \emph{explaining away effects} by the language model.
This primarily entails assigning low scores to unrelated outputs, which may not require a very powerful model.
In this case, we may be able to substantially decrease the size of the channel model at only a small loss in accuracy.

Recent work demonstrates that direct models with shallow decoders can give comparable accuracy, while being faster at inference time, compared to models with deep decoders~\citep{wu2019pay,elbayad2019depthadaptive,kasai2020deep,fan2019reducing}. 
This is particularly attractive for direct models for which the decoder network accounts for most of the wall time during inference but the dynamics for channel models are different:
the channel model repeatedly scores the entire input sequence given progressively larger target prefixes. 
Unlike for direct models, there is no straightforward way to reuse the encoder output between time-steps, and we opt to recompute the entire encoder and decoder of the channel model at every target time-step.
Since the input sequence is given, channel model computation can be batched over all tokens in the target prefix and the input sequence. This implies that we are free to adjust both the encoder and decoder depth.

We pursue two strategies to reduce model size: 
first, we progressively reduce the model dimension of the \cmbase{} Transformer architecture, by first halving the model dimension from 512 to 256, as well as the feed forward dimension from 2048 to 1024 for the \cmhalf{} model. 
The smallest configuration uses a model dimension of just 32 and a feed forward dimension of 128 (denoted as \cmsixteenth{} model).
Second, we consider models with only a single encoder block and a single decoder block. 
These models have a postfix \texttt{\_1\_1}, e.g., \cmsixteenths{}.
\autoref{tab:channel_sizes} shows the various model sizes as well as accuracy on the development set, newstest2016.

\begin{table}
\centering
\begin{tabular}{lcc}
\toprule
& Parameters (M) & BLEU \\
\midrule
big & 282.7 & 38.0 \\
big\_1\_1 & 93.8 & 34.1 \\
base & 72.1 & 36.7 \\
base\_1\_1 & 23.6 & 31.2 \\
half & 25.1 & 33.6 \\
half\_1\_1 & 15.8 & 27.4 \\
quarter & 9.8 & 28.4 \\
quarter\_1\_1 & 7.5 & 22.0 \\
16th & 2.8  & 15.9 \\
16th\_1\_1 & 2.7 & 10.0 \\
\bottomrule
\end{tabular}
\caption{Smaller channel models in terms of number of total parameters as well as BLEU (avg. over 3 seeds) on the development set. All models have six blocks each in the encoder and the decoder, except for models ending in "\_1\_1" which have only a single block in the encoder and the decoder.}
\label{tab:channel_sizes}
\end{table}

\subsection{Reducing the Output Vocabulary}
\label{sec:smallvocab}

During online noisy channel decoding, we need to allocate memory for a large number of channel model output probabilites ($k_1 \times k_2 \times S \times V$, as explained in \autoref{sec:ncapproach}). 
This substantially reduces the maximum possible batch size in order to prevent running out of memory while decoding on GPUs. 
A small batch size prevents the full utilization of parallel computation on GPUs, particularly, when the channel model is relatively small: some of our channel models have an embedding dimension of just 32.

To address this issue, we make use of the fact that we know exactly which input tokens need to be scored (since the input sequence is given) instead of computing probabilities for the entire vocabulary.
This is similar to vocabulary reduction techniques used for early neural sequence to sequence models, and it is particularly convenient since we know exactly which tokens are in the input sequence~\citep{mi2016vocabulary,hostis2016vocabulary}.

Similar to prior work on vocabulary reduction, we found it useful to not just score the input words but also a subset of the most frequent words in the vocabulary. 
Specifically, for each batch, we enumerate all input word types, add the 500 most frequent types and then compute output probabilities for this subset with the channel model. 
The number of output probabilities calculated is typically at least one order of magnitude smaller than the full vocabulary, as shown in \autoref{sec:ablation_small_output_vocab}.

This approach substantially reduces the memory footprint of small channel models and enables the use of much larger batch sizes which leads to faster inference as we will see in \autoref{sec:results}.

\subsection{Reducing the Number of Candidates}

We also study the effect of reducing the number of next token candidates $k_2$ scored for each beam at each step of beam search. 
This reduces the computation as well as memory overhead of channel model scoring.

\section{Experimental Setup}

\subsection{Datasets}
\label{sec:datasets}

We consider two datasets for our experiments:
For German-English (De-En), we train on WMT'19 training data. Following \citep{ng2019fairwmt}, we apply language identification filtering \citep{lui2012langid} and remove sentences longer than 250 tokens as well as sentence pairs with a source/target
length ratio exceeding 1.5. This results in 26.8M sentence pairs. We validate on newstest2016 and test on newstest2014, newstest2015, newstest2017, and newstest2018. 
For all models, the source vocabulary is a 24K byte pair encoding (BPE; Sennrich et al., 2016)\nocite{sennrich2016bpe} learned on the source portion of the bitext. 
For the target side, we use the vocabulary of the language model (\textsection\ref{sec:setup_lm}) so that both models score the exact same units during beam search.

For Romanian-English (Ro-En), we train on WMT'16 training data, comprising 612K sentence pairs, validate on newsdev2016 and test on newstest2016. 
We learn a joint BPE vocabulary of 18K types on the bitext training data which is used for both the source and target. 
Different to German-English, we learn a joint BPE vocabulary to enable sharing the source and target embeddings which we found to perform better for Romanian-English in early experiments.

\subsection{Language Models}
\label{sec:setup_lm}

For German-English, we train a sentence-level English Transformer language model with 16 layers and Transformer-Big architecture~\citep{vaswani2017attention,radford2018improving}.
The model is trained on de-duplicated English Newscrawl data from 2007-2018 comprising 186 million sentences or 4.5B words after normalization and tokenization. 
We use a BPE vocabulary of 24K types learned on this data.
For Romanian-English translation, we train a similar English Transformer language model that uses the joint BPE vocabulary learned on the Romanian-English bitext.
The latter enables the LM to score the exact same units as the sequence to sequence model during beam search.

We train a sentence-level Romanian Transformer language model with 16 layers and Transformer-Big architecture. The model is trained on de-duplicated Romanian CommonCrawl data consisting of 623M sentences or 21.7B words after normalization and tokenization~\citep{conneau2019xlmr, wenzek2020ccnet}.

The German-English bitext training data as well as the language model training data are preprocessed with the Moses tokenizer~\citep{koehn2007moses}. We normalize punctuation and remove non-printing characters.
Romanian-English data is pre-processed following~\citet{sennrich2016edinburgh} by applying Moses tokenization and special normalization for Romanian text.\footnote{\url{ https://github.com/rsennrich/wmt16-scripts/tree/master/preprocess}}

\subsection{Translation Models}

For De-En, we use the Transformer-Big architecture for the direct model. 
We do not share encoder and decoder embeddings since the source and target vocabularies are different. 
For channel models, operating from English to German, we consider different variants (\textsection\ref{sec:smallch}, \autoref{tab:channel_sizes}) to better understand the speed-accuracy trade-off of decreasing the capacity of channel models. 

For Ro-En and En-Ro with bitext only, the direct and channel models use a Transformer-Base architecture. For Ro-En with backtranslation, the direct and channel models use a Transformer-Big architecture. We share the encoder and decoder embeddings since the source and target vocabularies are the same and because this improved accuracy.

\subsection{Online Noisy Channel Decoding Setup}
\label{sec:setup}

In order to set weights for the linear combination of model scores (\autoref{eq:comb}), we randomly sample a set of hyperparameters and evaluate each configuration on the development set~\citep{yee2019noisy,ng2019fairwmt}. Hyperparameters are sampled within the interval $[0,2]$, 
For direct models (\dir{}), we sample ten random weights for the length penalty. 
For direct models combined with language models (\dirl{}), we evaluate 100 randomly sampled configurations for the length penalty and the language model weight ($\lambda_2$). 
For direct models combined with language models and channel models (\dirlc{}), we evaluate 1000 configurations of the length penalty, the language model weight ($\lambda_2$) and the channel model weight ($\lambda_1$). 
We use 16-bit floating point precision ~\citep{ott2018scaling,ott2019fairseq} for decoding with the online noisy channel setup.

Accuracy is measured via sacreBLEU \citep{post2018call} for WMT German-English. We report the average BLEU of the newstest2014-2015 and newstest2017-2018 test sets, averaged over 3 random seeds for model weight initialization. 
Speed is measured by the generation time (averaged over 3 trials) in seconds on the German-English newstest2016 test set on a 32GB Volta V100 GPU using 16-bit floating point precision~\citep{ott2018scaling,ott2019fairseq}. 
Unless otherwise specified, the beam size is 5, and the number of candidates for noisy channel model scoring per beam is $k_2=10$, unless otherwise specified. 
Generation times are based on a tuned batch size for each model configuration. We select the batch size within $(1, 10, 25, 50, 75, 100, 125, 150, 200, 300)$ that fits in memory and results in the fastest generation time.

\section{Results}
\label{sec:results}
\subsection{Fast Noisy Channel Modeling}

\begin{table}[t]
\centering
\begin{adjustbox}{width=\columnwidth}
\begin{tabular}{@{}lccc@{}}
\toprule
 &
 \multicolumn{1}{l}{\begin{tabular}[c]{@{}c@{}}Total\\Params\\ (M)\end{tabular}} &
 BLEU &
 Time (s) \\ \midrule
\multicolumn{3}{@{}l}{\textit{Ensembles}} \\
dir                        & \phantom{0}283 & 38.8 & \phantom{0}20\\
2 dir                      & \phantom{0}565 & 39.3 & \phantom{0}40 \\
3 dir                      & \phantom{0}848 & 39.5 & \phantom{0}59 \\
\midrule
\multicolumn{3}{@{}l}{\textit{Ensembles + LMs}} \\
dir + lm                   & \phantom{0}539 & 39.7 & \phantom{0}44 \\
2 dir + lm                 & \phantom{0}822 & 40.2 & \phantom{0}65 \\
3 dir + lm                 & 1104 & 40.3 & \phantom{0}84 \\
\midrule
\multicolumn{3}{@{}l}{\textit{Noisy Channel Modeling \citep{yee2019noisy}}} \\
dir + lm + big & \phantom{0}822 & 40.5 & 550 \\
\midrule
\multicolumn{3}{@{}l}{\textit{Fast Noisy Channel Modeling (This work)}} \\
dir + lm + 16th\_1\_1      & \phantom{0}542 & 40.2 & \phantom{0}56\\ 
dir + lm + base\_1\_1      & \phantom{0}574 & 40.5 & \phantom{0}92 \\
2 dir + lm + 16th\_1\_1    & \phantom{0}824 & 40.5 & \phantom{0}76 \\
2 dir + lm + base\_1\_1    & \phantom{0}857 & 40.8 & 111 \\
3 dir + lm + 16th\_1\_1    & 1107 & 40.6 & \phantom{0}93 \\
3 dir + lm + base\_1\_1    & 1140 & 41.0 & 131 \\
\bottomrule
\end{tabular}
\end{adjustbox}
\caption{Fast noisy channel modeling is more accurate than ensembles at comparable speed and the two methods are additive.
All results use beam size 5, batch sizes for each configuration are optimized and BLEU is averaged over news2014, news2015, news2017 and news2018 of WMT German to English.}
\label{table:ensemble_table}
\end{table}

\begin{table}[t]
\centering
\begin{adjustbox}{width=\columnwidth}
\begin{tabular}{@{}lccc@{}}
\toprule
 &\multicolumn{1}{l}{\begin{tabular}[c]{@{}c@{}}Channel\\Model\\Params\\(M)\end{tabular}} & 
 BLEU &
 Time (s) \\ \midrule
dir + lm + big             & 283 & 40.3 & 472 \\
dir + lm + base            & \phantom{0}72 & 40.4 & 202 \\
dir + lm + half            & \phantom{0}25 & 40.5 & 132 \\
dir + lm + quarter         & \phantom{0}10 & 40.4 & 102 \\
dir + lm + 8th             & \phantom{00}6 & 40.3 & \phantom{0}89 \\
dir + lm + 16th            & \phantom{00}3 & 40.2 & \phantom{0}70 \\
\midrule
dir + lm + big\_1\_1       & \phantom{0}94 & 40.5 & 160 \\
dir + lm + base\_1\_1      & \phantom{0}24 & 40.5 & \phantom{0}92 \\
dir + lm + half\_1\_1      & \phantom{0}16 & 40.4 & \phantom{0}72 \\
dir + lm + quarter\_1\_1   & \phantom{00}8 & 40.2 & \phantom{0}63 \\
dir + lm + 8th\_1\_1       & \phantom{00}5 & 40.3 & \phantom{0}60 \\
dir + lm + 16th\_1\_1      & \phantom{00}3 & 40.2 & \phantom{0}56 \\ 
\bottomrule
\end{tabular}
\end{adjustbox}
\caption{
Smaller channel models perform similarly for the standard beam size of 5. We exploit this fact to speed up noisy channel decoding. 
}

\label{table:small_channel_table}
\end{table}

In the first experiment, we evaluate the speed and accuracy of fast noisy channel decoding (\autoref{sec:fastnc}) and compare to the na\"{i}ve version without approximations \citep{yee2019noisy}.
As additional baselines, we consider a single direct model (\dir{}), ensembling two direct models (\ddir{}) and three direct models (\dddir{}), as well as adding a language model to each (\texttt{lm}).
As channel models, we consider a \cmbig{} Transformer, a \cmbase{} Transformer, as well as a variant with model dimension of only 32 which is 1/16th of the model dimension of a \cmbase{} Transformer with a single layer in the encoder and decoder each (\cmsixteenths{}), totaling just 2.7M parameters.
For fast noisy channel decoding, we reduce the channel model output vocabulary (\textsection\ref{sec:smallvocab}) and set $k_2=3$; we ablate these choices in \autoref{sec:ablations}.

\autoref{table:ensemble_table} shows that the approximations we introduce to make noisy channel decoding fast also achieve similar accuracy (40.5 BLEU) to the much slower noisy channel approach of~\citep{yee2019noisy}, while being about six times faster at inference time.

\autoref{table:ensemble_table} also shows that \texttt{dir + lm + 16th\_1\_1} is 0.7 BLEU score better than \dddir{} at a similar decoding speed. Thus, using a small channel model and a language model with online noisy channel decoding is a better strategy than ensembling 3 direct models.
Noisy channel decoding is also complementary to ensembling direct models: \texttt{3 dir + lm + base\_1\_1} improves by 0.7 BLEU compared to \dddirl{}.

\autoref{table:small_channel_table} compares fast noisy channel decoding with different channel model sizes.
Generally, smaller channel models are only slightly less accurate than larger models while being significantly faster than their larger counterparts.
For example, \texttt{16th\_1\_1} is over eight times faster than \texttt{big} and achieves nearly the same accuracy.

This observation is in line with the hypothesis that the primary role of the channel model is to tie back the language model generations to the input.
We exploit the fact that small channel models work well to make noisy channel decoding very fast.

\subsection{Noisy Channel Decoding with Larger Beam Sizes}

\begin{figure}[t]
\centering
\begin{tikzpicture}
\pgfplotsset{
    every axis legend/.append style={
        at={(0.4,1.05)},
        anchor= south
    },
    grid style={lightgray!60!white}
}
\begin{axis}[
    height=1.15\columnwidth,
    width=\columnwidth,
    legend style={font=\footnotesize},
    legend cell align={left},
    legend columns=3,
    xmode=linear,
    xlabel=Beam Size,
    ylabel=BLEU,
    xtick={5,10,25,50},
    xticklabels={5,10,25,50},
    ytick={38.6,38.8,39.0,39.2,39.4,39.6,39.8,40.0,40.2,40.4,40.6,40.8,41.0},
    yticklabels={38.6,,39.0,,39.4,,39.8,,40.2,,40.6,,41.0},
    ytick distance=0.2,
    ymajorgrids,
];
\addplot[violet!70!white, mark=x, style={thick}] table [y=dir, x=type]{data/bleu_vs_beam.dat};
\addplot[yellow!70!red, mark=o, style={thick}] table [y=2dir, x=type]{data/bleu_vs_beam.dat};
\addplot[darkgray, mark=asterisk, style={thick}] table [y=3dir, x=type]{data/bleu_vs_beam.dat};
\addplot[red, mark=*, style={thick}] table [y=big, x=type]{data/bleu_vs_beam.dat};
\addplot[blue, mark=triangle, style={thick}] table [y=big11, x=type]{data/bleu_vs_beam.dat};
\addplot[green!80!black, mark=square, style={thick}] table [y=base, x=type]{data/bleu_vs_beam.dat};
\addplot[cyan!90!black, mark=+, style={thick}] table [y=quarter11, x=type]{data/bleu_vs_beam.dat};
\addplot[magenta!90!white, mark=diamond, style={thick}] table [y=16th11, x=type]{data/bleu_vs_beam.dat};

\legend{dir,2 dir,3 dir,fast big,fast big\_1\_1,fast base,fast quarter\_1\_1,fast 16th\_1\_1}
\end{axis}
\end{tikzpicture}
\caption{BLEU of fast online noisy channel decoding with different channel models when beam size is increased (compared to ensemble baselines). BLEU is averaged over news2014, news2015, news2017 and news2018 of WMT De-En. 
}
\label{fig:blue_vs_beam_src_vocab}
\end{figure}
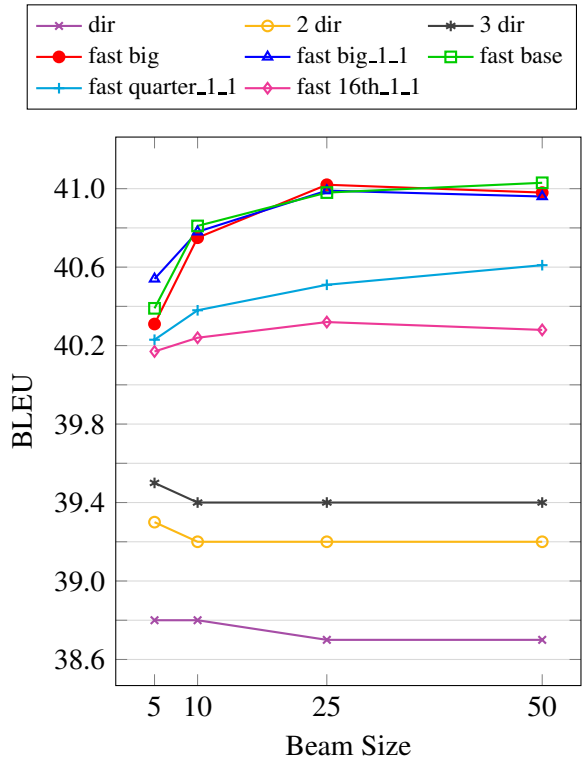

So far we used a standard beam size of five to enable fast decoding. 
However, previous work found that noisy channel modeling benefits more from larger beam sizes than other methods~\citep{yee2019noisy}.
Next, we evaluate whether our efficiency improvements still enable good performance with larger beam sizes.

\autoref{fig:blue_vs_beam_src_vocab} shows that for beam size 5, most channel models perform comparably. 
Larger models are slightly better but overall they are in a similar ball park.
As the beam size increases, larger channel models do achieve better accuracy.
However, there is no difference between a single layer big model (\cmbigs{}) and a six layer version (\cmbig{}).
As observed in previous work~\cite{yee2019noisy}, the direct model and the direct ensembles (\dir{}, \ddir{}, \dddir{}) do not benefit from larger beam sizes.

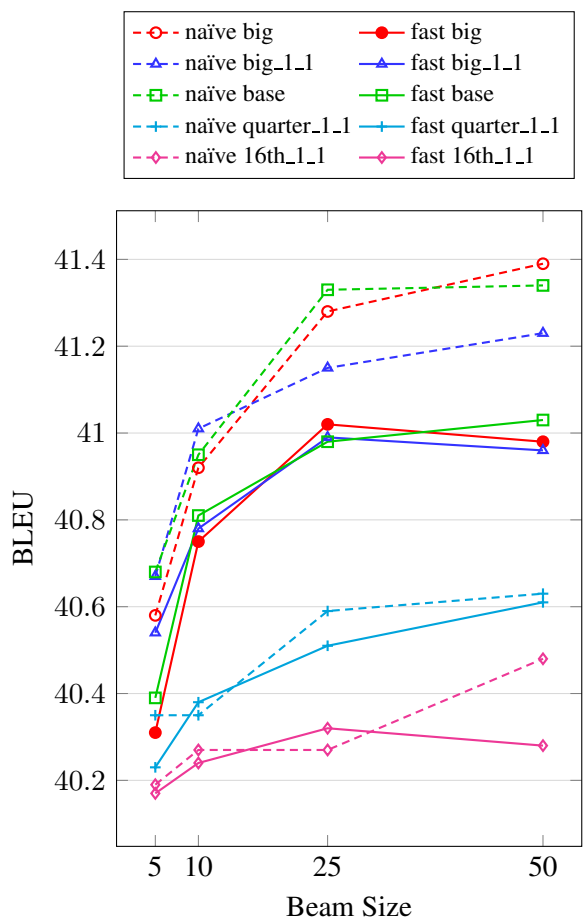
\begin{figure}[t]
\centering
\begin{tikzpicture}
\pgfplotsset{
    every axis legend/.append style={
        at={(0.5,1.05)},
        anchor= south
    },
    grid style={lightgray!60!white}
}
\begin{axis}[
    height=1.3\columnwidth,
    width=\columnwidth,
    legend style={font=\footnotesize},
    legend cell align={left},
    legend columns=2,
    xmode=linear,
    xlabel=Beam Size,
    ylabel=BLEU,
    xtick={5,10,25,50},
    xticklabels={5,10,25,50},
    ytick distance=0.2,
    ymajorgrids,
];
\addplot[red, mark=o, densely dashed, mark options=solid, style={thick}] table [y=slowbig, x=type]{data/bleu_vs_beam.dat};
\addplot[red, mark=*, style={thick}] table [y=big, x=type]{data/bleu_vs_beam.dat};
\addplot[blue!80!white, mark=triangle, densely dashed, mark options=solid, style={thick}] table [y=slowbig11, x=type]{data/bleu_vs_beam.dat};
\addplot[blue!80!white, mark=triangle, style={thick}] table [y=big11, x=type]{data/bleu_vs_beam.dat};
\addplot[green!80!black, mark=square, densely dashed, mark options=solid, style={thick}] table [y=slowbase, x=type]{data/bleu_vs_beam.dat};
\addplot[green!80!black, mark=square, style={thick}] table [y=base, x=type]{data/bleu_vs_beam.dat};
\addplot[cyan!90!black, mark=+, densely dashed, mark options=solid, style={thick}] table [y=slowquarter11, x=type]{data/bleu_vs_beam.dat};
\addplot[cyan!90!black, mark=+, style={thick}] table [y=quarter11, x=type]{data/bleu_vs_beam.dat};
\addplot[magenta!90!white, mark=diamond, densely dashed, mark options=solid, style={thick}] table [y=slow16th11, x=type]{data/bleu_vs_beam.dat};
\addplot[magenta!90!white, mark=diamond, style={thick}] table [y=16th11, x=type]{data/bleu_vs_beam.dat};

\legend{na\"{i}ve big,fast big,na\"{i}ve big\_1\_1,fast big\_1\_1,na\"{i}ve base,fast base,na\"{i}ve quarter\_1\_1,fast quarter\_1\_1,na\"{i}ve 16th\_1\_1,fast 16th\_1\_1}
\end{axis}
\end{tikzpicture}
\caption{BLEU of fast and na\"{i}ve online noisy channel decoding with different channel models sizes when beam size is increased. BLEU is averaged over news2014, news2015, news2017 and news2018 of WMT De-En.
}
\label{fig:blue_vs_beam_slow_fast}
\end{figure}

\begin{figure}[ht]
\centering
\begin{adjustbox}{width=\columnwidth}
\begin{tikzpicture}
\pgfplotsset{
    grid style={lightgray!60!white}
}
\begin{axis}[
    legend style={font=\footnotesize},
    legend cell align={left},
    xmode=linear,
    xtick={5,10,25,50},
    xticklabels={5,10,25,50},
    ytick={100,200,500,1000,3000},
    yticklabels={100,200,500,1000,3000},
    xlabel=Beam Size,
    ylabel=Generation Time (secs) (log scale),
    ymode=log,
    legend style={at={(0.6, 0.22)}, anchor=near xticklabel},
    ymajorgrids
];
\addplot[blue, mark=o] table [y=fast, x=type]{data/big_1_1_speed_fast_vs_slow_beams.dat};
\addplot[red, mark=x] table [y=naive, x=type]{data/big_1_1_speed_fast_vs_slow_beams.dat};
\addlegendentry{Fast Noisy Channel Decoding};
\addlegendentry{Na\"{i}ve Noisy Channel Decoding};
\end{axis}
\end{tikzpicture}
\end{adjustbox}
\caption{With larger beam sizes, the speed of fast approximations for noisy channel decoding scales much better than that of na\"{i}ve noisy channel decoding. Results are based on generation using the \cmbigs{} channel model with the fastest batch size for each setting with beam 5 on newstest2016 De-En.}
\label{fig:gen_time_slow_vs_fast_beams}
\end{figure}
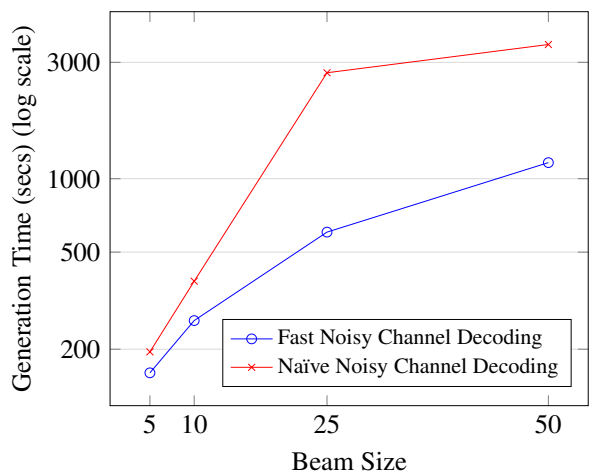

Next, we compare fast noisy channel decoding and na\"{i}ve noisy channel decoding at larger beam sizes. As shown in \autoref{fig:gen_time_slow_vs_fast_beams}, the na\"{i}ve approach is much slower. Fast approximations to noisy channel decoding scale much better in terms of speed as the beam size increases. \autoref{fig:blue_vs_beam_slow_fast} compares the accuracy of fast noisy channel decoding at larger beam sizes with that of na\"{i}ve noisy channel decoding. Using the \cmbig{} and \cmbigs{} channel models gives the best performance across all beam sizes for na\"{i}ve noisy channel decoding.
With fast noisy channel decoding, we see an average drop of 0.3 BLEU and 0.2 BLEU for \cmbig{} and \cmbigs{} respectively. 
On the other hand, for smaller channel models, the difference between na\"{i}ve and fast noisy channel decoding is generally smaller.

\subsection{Results on WMT Romanian-English}

\begin{table*}
\centering
\begin{tabular}{@{}llllll@{}}
\toprule
 &
 \begin{tabular}[c]{@{}l@{}}mono\\tokens\\Ro-En\\(B)\end{tabular} & 
 \begin{tabular}[c]{@{}l@{}}mono\\tokens\\En-Ro\\(B)\end{tabular} &
  En-Ro &
  Ro-En &
  \begin{tabular}[c]{@{}l@{}}Ro-En\\+BT\end{tabular} \\ \midrule
mBART02  & 66 & 66 & 38.5 & 38.5 & 39.9 \\
mBART02 (beam=50) & 66 & 66 & - & - & 39.9 \\ \midrule
dir       & - & - & 34.6 & 34.6 & 38.4 \\
dir + lm    & 4.5 & 22 & 35.4 & 35.9 & 38.7 \\
dir + lm + big & 4.5 & 22 & 37.3 & 37.7 & 39.6 \\
dir + lm + big (beam=50) &
    4.5 & 22 & 
  \textbf{39.1} &
  \textbf{39.1} &
  \textbf{40.4} \\ \bottomrule
\end{tabular}
\caption{BLEU of noisy channel decoding on the Romanian-English newstest2016 test set with bitext-only as well as with backtranslation (BT) compared to mBART~\citep{liu2020multilingual}. 
We also show the total amount of monolingual data used by each method in billions of tokens.  
}
\label{table:ro_en}
\end{table*}

\begin{table}[t]
\centering
\begin{adjustbox}{width=\columnwidth}
\begin{tabular}{@{}lcr@{}}
\toprule
 &
 BLEU &
 Time (s) \\ \midrule
mBART02                        & 39.9 & 93 \\
mBART02 (beam=50)              & 39.9 & 754  \\
\midrule
dir                            & 38.4 & 19 \\
\midrule
\multicolumn{3}{@{}l}{\textit{Noisy Channel Modeling \citep{yee2019noisy}}} \\
dir + lm + big               & 39.6 & 1178 \\
dir + lm + big (beam=50)     & 40.4 & 12554 \\
\midrule
\multicolumn{3}{@{}l}{\textit{Fast Noisy Channel Modeling}} \\
dir + lm + base\_1\_1                 & 39.8 & 82 \\
dir + lm + base\_1\_1 (beam=50)       & 40.3 & 631 \\
\bottomrule
\end{tabular}
\end{adjustbox}
\caption{Speed and accuracy on Romanian-English (Ro-En) with backtranslation. 
Fast noisy channel decoding using \cmbases{} achieves similar accuracy to mBART02 while being faster (beam=5). 
BLEU is measured on newstest2016 and generation time is measured on newsdev2016. }
\label{table:ro_en_fast}
\end{table}

Next, we evaluate noisy channel modeling on WMT Romanian-English translation (Ro-En and En-Ro) which is a low resource setup  compared to WMT German-English. 
We also compare to a recently introduced pre-training approach, mBART. The mBART model is pre-trained to denoise input sentences in multiple languages, followed by fine-tuning on the bitext~\citep{liu2020multilingual}.
Following their setup for En-Ro evaluation, we apply Moses tokenization and normalize diacritics for Romanian \citep{sennrich2016edinburgh}, and use tokenized BLEU. 
For Ro-En, we use SacreBLEU~\citep{post2018call}.

\autoref{table:ro_en} shows that noisy channel decoding with a wide beam can outperform multilingual pre-training (mBART) across the board.
Large beams are not helpful for generation with mBART.
Compared to the direct model, noisy channel decoding improves by 2.7/3.1 BLEU on En-Ro and Ro-En respectively, and increasing the beam size gives gains of 4.5/4.5 BLEU.

We also study the performance of noisy channel decoding on Romanian-English with backtranslated data generated using unrestricted sampling~\citep{edunov2018bt}.\footnote{The monolingual English data used for backtranslation comes from \url{http://data.statmt.org/rsennrich/wmt16\_backtranslations/} \citep{sennrich2016improving}.}
As compared to mBART02~\citep{liu2020multilingual}, the previous state-of-the-art result on Romanian-English with backtranslation, we achieve a 0.5 BLEU improvement. 
We use a similar number of total model parameters, but much less monolingual English data. 
Our English language model is trained on 4.5B tokens, while mBART02 uses 66B tokens of English and Romanian monolingual data.

Finally,~\autoref{table:ro_en_fast} shows that fast approximations and smaller channel models achieve similar performance but much higher speed compared to na\"{i}ve noisy channel decoding on WMT Romanian-English with back-translation.
Fast noisy channel decoding with \cmbases{} achieves comparable accuracy as mBART02 at slightly faster generation time with beam size 5.

\subsection{Ablations}
\label{sec:ablations}

In this section we focus on some of the design choices we made to speed up noisy channel decoding.
We measure the impact on speed and accuracy when reducing the output vocabulary size of the channel model, and reducing the number of beam candidates scored by the channel model.

\subsubsection{Reducing the Output Vocabulary}
\label{sec:ablation_small_output_vocab}

\begin{table}[t]
\centering
\begin{adjustbox}{width=\columnwidth}
\begin{tabular}{@{}lccccc@{}}
\toprule
\multicolumn{1}{c}{\begin{tabular}[c]{@{}c@{}}\textbf{dir+ch+lm}\\\textbf{(beam=5)}\end{tabular}} &
  \multicolumn{2}{c}{\begin{tabular}[c]{@{}c@{}}Full Source\\ Vocab\end{tabular}} &&
  \multicolumn{2}{c}{\begin{tabular}[c]{@{}c@{}}Small Source\\Vocab\end{tabular}} \\
  \cmidrule{2-3}
  \cmidrule{5-6}
 &
  BLEU &
  \begin{tabular}[c]{@{}c@{}}Time (s)\end{tabular} &&
  BLEU &
  \begin{tabular}[c]{@{}c@{}}Time (s)\end{tabular} \\ \midrule
big           & 40.6 &           1656 && 40.3 &           1355 \\
base          & 40.7 & \phantom{0}854 && 40.4 & \phantom{0}516 \\
half          & 40.6 & \phantom{0}450 && 40.5 & \phantom{0}299 \\
quarter       & 40.5 & \phantom{0}359 && 40.4 & \phantom{0}212 \\
8th           & 40.3 & \phantom{0}324 && 40.3 & \phantom{0}178 \\
16th          & 40.1 & \phantom{0}264 && 40.2 & \phantom{0}118 \\
\midrule
big\_1\_1     & 40.7 & \phantom{0}543 && 40.5 & \phantom{0}339 \\
base\_1\_1    & 40.5 & \phantom{0}336 && 40.5 & \phantom{0}169 \\
half\_1\_1    & 40.3 & \phantom{0}264 && 40.4 & \phantom{0}117 \\
quarter\_1\_1 & 40.4 & \phantom{0}238 && 40.2 & \phantom{00}95 \\
8th\_1\_1     & 40.1 & \phantom{0}223 && 40.3 & \phantom{00}87 \\
16th\_1\_1    & 40.2 & \phantom{0}209 && 40.2 & \phantom{00}74 \\
\bottomrule
\end{tabular}
\end{adjustbox}
\caption{Comparison of accuracy (BLEU) and speed of online noisy channel decoding with and without the small output vocabulary approximation for different channel model sizes. 
Note we use $k_2=10$ for this ablation. BLEU is averaged over news2014, news2015, news2017 and news2018 of WMT De-En and generation time is on news2016.
}
\label{tab:src_vocab}
\end{table}

In the next experiment, we compare the speed of using the full output vocabulary for the channel model to a reduced version. 
Specifically, we reduce the vocabulary by selecting all source tokens in the batch as well as the most frequent 500 tokens in the training data (see~\autoref{sec:smallvocab}).
We tune each setup by selecting the fastest batch size based on a sweep over different batch sizes $(1, 10, 25, 50, 75, 100, 125, 150, 200, 300)$.

\autoref{tab:src_vocab} shows that generating channel model scores for a small subset of the source vocabulary results in a small accuracy of up to 0.3 BLEU, but often less, while substantially increasing speed by 40-65\% for single layer channel models and by 20-55\% for other channel models.
\cmbases{} with a small vocabulary is nearly ten times faster than the approach proposed in~\citet{yee2019noisy} (channel model size \cmbig{}), with a slight decline in accuracy.

The average vocabulary size used for scoring the channel model is around 1050, as compared to full source vocabulary size of 28,048. 
This leads to a large reduction in memory consumption and enables fitting larger batches into memory.



\subsubsection{Reducing the Number of Candidates}

\begin{table}
\centering
\begin{tabular}{@{}lccc@{}}
\toprule
& $k_2$  & BLEU & Time (s) \\ \midrule
& 2  & 40.4   &  \phantom{0}76            \\
& 3  & 40.5   &  \phantom{0}88            \\
& 5  & 40.4   & 124           \\
& 10 & 40.5   & 168           \\ \bottomrule
\end{tabular}
\caption{Smaller number of rescoring candidates $k_2$ per beam are as accurate and much faster than larger values of $k_2$ for fast noisy channel decoding using \cmbases{} with beam 5. BLEU is averaged over news2014, news2015, news2017 and news2018 of WMT De-En and generation time is on news2016.
}
\label{tab:k-values}
\end{table}
For each beam in each step of beam search, we need to make a choice about how many candidates $k_2$ we re-score with noisy channel modeling. \citet{yee2019noisy} re-scored $k_{2}=10$ candidates for each beam at each step. We sweep over different values of $k_2$ to understand the speed-accuracy trade-off associated with the choice of $k_2$.
Table \ref{tab:k-values} shows that smaller values for $k_2$ are as accurate and much faster for beam size 5.

\section{Conclusion}

We introduced a number of approximations which greatly speed up noisy channel modeling for neural sequence to sequence models. 
This includes using channel models which are a fraction of the size of commonly used sequence to sequence models, pruning most of the channel model output vocabulary, and reducing the number of beam candidates scored by the channel model.

Our approximations are simple, yet, highly effective and enable comparable inference speed to ensembles of direct models while delivering higher accuracy.
Our experiments show that noisy channel modeling can outperform pre-training approaches by being able to better exploit wider beams.
Moreover, this is achieved while using a smaller amount of monolingual data.

\bibliography{anthology,master}

\begin{thebibliography}{37}
\expandafter\ifx\csname natexlab\endcsname\relax\def\natexlab#1{#1}\fi

\bibitem[{Bojar and Tamchyna(2011)}]{bojar2011bt}
Ondrej Bojar and Ales Tamchyna. 2011.
\newblock Improving translation model by monolingual data.
\newblock In \emph{Proc. of WMT}.

\bibitem[{Brown et~al.(1993)Brown, Pietra, Pietra, and
  Mercer}]{brown1993mathematics}
Peter~F Brown, Vincent J~Della Pietra, Stephen A~Della Pietra, and Robert~L
  Mercer. 1993.
\newblock The mathematics of statistical machine translation: Parameter
  estimation.
\newblock \emph{Computational linguistics}.

\bibitem[{Chen et~al.(2019)Chen, Shen, Le, Chaudhary, El-Kishky, Wenzek, Ott,
  and Ranzato}]{chen2019facebook}
Peng-Jen Chen, Jiajun Shen, Matt Le, Vishrav Chaudhary, Ahmed El-Kishky,
  Guillaume Wenzek, Myle Ott, and Marc’Aurelio Ranzato. 2019.
\newblock Facebook ai’s wat19 myanmar-english translation task submission.
\newblock \emph{WAT 2019}, page 112.

\bibitem[{Conneau et~al.(2019)Conneau, Khandelwal, Goyal, Chaudhary, Wenzek,
  Guzmán, Grave, Ott, Zettlemoyer, and Stoyanov}]{conneau2019xlmr}
Alexis Conneau, Kartikay Khandelwal, Naman Goyal, Vishrav Chaudhary, Guillaume
  Wenzek, Francisco Guzmán, Edouard Grave, Myle Ott, Luke Zettlemoyer, and
  Veselin Stoyanov. 2019.
\newblock Unsupervised cross-lingual representation learning at scale.
\newblock \emph{arXiv}.

\bibitem[{Devlin et~al.(2019)Devlin, Chang, Lee, and
  Toutanova}]{devlin2019bert}
Jacob Devlin, Ming-Wei Chang, Kenton Lee, and Kristina Toutanova. 2019.
\newblock Bert: Pre-training of deep bidirectional transformers for language
  understanding.
\newblock In \emph{Proc. of NAACL}.

\bibitem[{Edunov et~al.(2019)Edunov, Baevski, and Auli}]{edunov2019lmpretrain}
Sergey Edunov, Alexei Baevski, and Michael Auli. 2019.
\newblock Pre-trained language model representations for language generation.
\newblock In \emph{Proc. of NAACL}.

\bibitem[{Edunov et~al.(2018)Edunov, Ott, Auli, and Grangier}]{edunov2018bt}
Sergey Edunov, Myle Ott, Michael Auli, and David Grangier. 2018.
\newblock Understanding back-translation at scale.
\newblock In \emph{Proc. of EMNLP}.

\bibitem[{Elbayad et~al.(2020)Elbayad, Gu, Grave, and
  Auli}]{elbayad2019depthadaptive}
Maha Elbayad, Jiatao Gu, Edouard Grave, and Michael Auli. 2020.
\newblock Depth-adaptive transformer.
\newblock In \emph{Proc. of ICLR}.

\bibitem[{Fan et~al.(2020)Fan, Grave, and Joulin}]{fan2019reducing}
Angela Fan, Edouard Grave, and Armand Joulin. 2020.
\newblock Reducing transformer depth on demand with structured dropout.
\newblock In \emph{Proc. of ICLR}.

\bibitem[{G{\"{u}}l{\c{c}}ehre et~al.(2015)G{\"{u}}l{\c{c}}ehre, Firat, Xu,
  Cho, Barrault, Lin, Bougares, Schwenk, and Bengio}]{gulcehre2015mono}
{\c{C}}aglar G{\"{u}}l{\c{c}}ehre, Orhan Firat, Kelvin Xu, Kyunghyun Cho,
  Lo{\"{\i}}c Barrault, Huei{-}Chi Lin, Fethi Bougares, Holger Schwenk, and
  Yoshua Bengio. 2015.
\newblock On using monolingual corpora in neural machine translation.
\newblock \emph{arXiv}, abs/1503.03535.

\bibitem[{He et~al.(2020)He, Gu, Shen, and Ranzato}]{he2019revisiting}
Junxian He, Jiatao Gu, Jiajun Shen, and Marc'Aurelio Ranzato. 2020.
\newblock Revisiting self-training for neural sequence generation.
\newblock In \emph{Proc. of ICLR}.

\bibitem[{Kasai et~al.(2020)Kasai, Pappas, Peng, Cross, and
  Smith}]{kasai2020deep}
Jungo Kasai, Nikolaos Pappas, Hao Peng, James Cross, and Noah~A. Smith. 2020.
\newblock Deep encoder, shallow decoder: Reevaluating the speed-quality
  tradeoff in machine translation.
\newblock \emph{arXiv}, abs/2006.10369.

\bibitem[{Klein and Manning(2001)}]{klein2001explain}
Dan Klein and Christopher Manning. 2001.
\newblock Conditional structure versus conditional estimation in nlp.
\newblock In \emph{Proc. of EMNLP}.

\bibitem[{Koehn et~al.(2007)Koehn, Hoang, Birch, Callison-Burch, Federico,
  Bertoldi, Cowan, Shen, Moran, Zens et~al.}]{koehn2007moses}
Philipp Koehn, Hieu Hoang, Alexandra Birch, Chris Callison-Burch, Marcello
  Federico, Nicola Bertoldi, Brooke Cowan, Wade Shen, Christine Moran, Richard
  Zens, et~al. 2007.
\newblock Moses: Open source toolkit for statistical machine translation.
\newblock In \emph{Proceedings of the 45th annual meeting of the ACL on
  interactive poster and demonstration sessions}, pages 177--180. Association
  for Computational Linguistics.

\bibitem[{Koehn et~al.(2003)Koehn, Och, and Marcu}]{koehn2003statistical}
Philipp Koehn, Franz~Josef Och, and Daniel Marcu. 2003.
\newblock Statistical phrase-based translation.
\newblock In \emph{Proc. of NAACL}.

\bibitem[{L'Hostis et~al.(2016)L'Hostis, Grangier, and
  Auli}]{hostis2016vocabulary}
Gurvan L'Hostis, David Grangier, and Michael Auli. 2016.
\newblock {Vocabulary Selection Strategies for Neural Machine Translation}.
\newblock \emph{arXiv}, abs/1610.00072.

\bibitem[{Li et~al.(2019)Li, Roller, Kulikov, Welleck, Boureau, Cho, and
  Weston}]{li2019dont}
Margaret Li, Stephen Roller, Ilia Kulikov, Sean Welleck, Y-Lan Boureau,
  Kyunghyun Cho, and Jason Weston. 2019.
\newblock Don't say that! making inconsistent dialogue unlikely with
  unlikelihood training.
\newblock \emph{arXiv}.

\bibitem[{Liu et~al.(2020)Liu, Gu, Goyal, Li, Edunov, Ghazvininejad, Lewis, and
  Zettlemoyer}]{liu2020multilingual}
Yinhan Liu, Jiatao Gu, Naman Goyal, Xian Li, Sergey Edunov, Marjan
  Ghazvininejad, Mike Lewis, and Luke Zettlemoyer. 2020.
\newblock Multilingual denoising pre-training for neural machine translation.
\newblock \emph{arXiv preprint arXiv:2001.08210}.

\bibitem[{Lui and Baldwin(2012)}]{lui2012langid}
Marco Lui and Timothy Baldwin. 2012.
\newblock langid. py: An off-the-shelf language identification tool.
\newblock In \emph{Proceedings of the ACL 2012 system demonstrations}, pages
  25--30.

\bibitem[{Mi et~al.(2016)Mi, Wang, and Ittycheriah}]{mi2016vocabulary}
Haitao Mi, Zhiguo Wang, and Abe Ittycheriah. 2016.
\newblock Vocabulary manipulation for neural machine translation.
\newblock In \emph{Proc. of ACL}.

\bibitem[{Ng et~al.(2019)Ng, Yee, Baevski, Ott, Auli, and
  Edunov}]{ng2019fairwmt}
Nathan Ng, Kyra Yee, Alexei Baevski, Myle Ott, Michael Auli, and Sergey Edunov.
  2019.
\newblock Facebook fair's wmt19 news translation task submission.
\newblock In \emph{Proc. of WMT}.

\bibitem[{Ott et~al.(2019)Ott, Edunov, Baevski, Fan, Gross, Ng, Grangier, and
  Auli}]{ott2019fairseq}
Myle Ott, Sergey Edunov, Alexei Baevski, Angela Fan, Sam Gross, Nathan Ng,
  David Grangier, and Michael Auli. 2019.
\newblock fairseq: A fast, extensible toolkit for sequence modeling.
\newblock In \emph{Proc. of NAACL System Demonstrations}.

\bibitem[{Ott et~al.(2018)Ott, Edunov, Grangier, and Auli}]{ott2018scaling}
Myle Ott, Sergey Edunov, David Grangier, and Michael Auli. 2018.
\newblock Scaling neural machine translation.
\newblock In \emph{Proceedings of the Third Conference on Machine Translation:
  Research Papers}, pages 1--9.

\bibitem[{Post(2018)}]{post2018call}
Matt Post. 2018.
\newblock A call for clarity in reporting bleu scores.
\newblock In \emph{Proceedings of the Third Conference on Machine Translation:
  Research Papers}, pages 186--191.

\bibitem[{Radford et~al.(2018)Radford, Narasimhan, Salimans, and
  Sutskever}]{radford2018improving}
Alec Radford, Karthik Narasimhan, Tim Salimans, and Ilya Sutskever. 2018.
\newblock Improving language understanding by generative pre-training.

\bibitem[{Raffel et~al.(2020)Raffel, Shazeer, Roberts, Lee, Narang, Matena,
  Zhou, Li, and Liu}]{raffel2020exploring}
Colin Raffel, Noam Shazeer, Adam Roberts, Katherine Lee, Sharan Narang, Michael
  Matena, Yanqi Zhou, Wei Li, and Peter~J Liu. 2020.
\newblock Exploring the limits of transfer learning with a unified text-to-text
  transformer.
\newblock \emph{Journal of Machine Learning Research}, 21(140):1--67.

\bibitem[{Sennrich et~al.(2016{\natexlab{a}})Sennrich, Haddow, and
  Birch}]{sennrich2016edinburgh}
Rico Sennrich, Barry Haddow, and Alexandra Birch. 2016{\natexlab{a}}.
\newblock Edinburgh neural machine translation systems for wmt 16.
\newblock In \emph{Proceedings of the First Conference on Machine Translation:
  Volume 2, Shared Task Papers}, pages 371--376.

\bibitem[{Sennrich et~al.(2016{\natexlab{b}})Sennrich, Haddow, and
  Birch}]{sennrich2016bt}
Rico Sennrich, Barry Haddow, and Alexandra Birch. 2016{\natexlab{b}}.
\newblock Improving neural machine translation models with monolingual data.
\newblock In \emph{Proc. of ACL}.

\bibitem[{Sennrich et~al.(2016{\natexlab{c}})Sennrich, Haddow, and
  Birch}]{sennrich2016improving}
Rico Sennrich, Barry Haddow, and Alexandra Birch. 2016{\natexlab{c}}.
\newblock Improving neural machine translation models with monolingual data.
\newblock In \emph{Proceedings of the 54th Annual Meeting of the Association
  for Computational Linguistics (Volume 1: Long Papers)}, pages 86--96.

\bibitem[{Sennrich et~al.(2016{\natexlab{d}})Sennrich, Haddow, and
  Birch}]{sennrich2016bpe}
Rico Sennrich, Barry Haddow, and Alexandra Birch. 2016{\natexlab{d}}.
\newblock Neural machine translation of rare words with subword units.
\newblock In \emph{Proc. of ACL}.

\bibitem[{Stahlberg et~al.(2018)Stahlberg, Cross, and
  Stoyanov}]{stahlberg2018simple}
Felix Stahlberg, James Cross, and Veselin Stoyanov. 2018.
\newblock Simple fusion: Return of the language model.
\newblock In \emph{Proc. of WMT}.

\bibitem[{Vaswani et~al.(2017)Vaswani, Shazeer, Parmar, Uszkoreit, Jones,
  Gomez, Kaiser, and Polosukhin}]{vaswani2017attention}
Ashish Vaswani, Noam Shazeer, Niki Parmar, Jakob Uszkoreit, Llion Jones,
  Aidan~N Gomez, {\L}ukasz Kaiser, and Illia Polosukhin. 2017.
\newblock Attention is all you need.
\newblock In \emph{Advances in neural information processing systems}, pages
  5998--6008.

\bibitem[{Wenzek et~al.(2020)Wenzek, Lachaux, Conneau, Chaudhary, Guzm{\'a}n,
  Joulin, and Grave}]{wenzek2020ccnet}
Guillaume Wenzek, Marie-Anne Lachaux, Alexis Conneau, Vishrav Chaudhary,
  Francisco Guzm{\'a}n, Armand Joulin, and {\'E}douard Grave. 2020.
\newblock Ccnet: Extracting high quality monolingual datasets from web crawl
  data.
\newblock In \emph{Proceedings of The 12th Language Resources and Evaluation
  Conference}, pages 4003--4012.

\bibitem[{Wu et~al.(2019)Wu, Fan, Baevski, Dauphin, and Auli}]{wu2019pay}
Felix Wu, Angela Fan, Alexei Baevski, Yann~N. Dauphin, and Michael Auli. 2019.
\newblock Pay less attention with lightweight and dynamic convolutions.
\newblock In \emph{Proc. of ICLR}.

\bibitem[{Yee et~al.(2019)Yee, Dauphin, and Auli}]{yee2019noisy}
Kyra Yee, Yann~N Dauphin, and Michael Auli. 2019.
\newblock Simple and effective noisy channel modeling for neural machine
  translation.
\newblock In \emph{Proc. of NAACL}.

\bibitem[{Yu et~al.(2017)Yu, Blunsom, Dyer, Grefenstette, and
  Kocisk{\'{y}}}]{yu2017neuralnoisy}
Lei Yu, Phil Blunsom, Chris Dyer, Edward Grefenstette, and Tom{\'{a}}s
  Kocisk{\'{y}}. 2017.
\newblock The neural noisy channel.
\newblock In \emph{Proc. of ICLR}.

\bibitem[{Yu et~al.(2020)Yu, Sartran, Stokowiec, Ling, Kong, Blunsom, and
  Dyer}]{yu2020tacl}
Lei Yu, Laurent Sartran, Wojciech Stokowiec, Wang Ling, Lingpeng Kong, Phil
  Blunsom, and Chris Dyer. 2020.
\newblock Better document-level machine translation with bayes’ rule.
\newblock \emph{TACL}.

\end{thebibliography}
\bibliographystyle{acl_natbib}

\clearpage
\appendix


\end{document}